\documentclass[runningheads]{llncs}
\usepackage{graphicx}
\usepackage{subfigure}
\usepackage{tikz}
\usepackage{comment}
\usepackage{amsmath,amssymb} 
\usepackage{color}
\usepackage{multirow}
\usepackage[misc,geometry]{ifsym}
\usepackage[accsupp]{axessibility}  
\usepackage[pagebackref,breaklinks,colorlinks]{hyperref}

\begin{document}
\pagestyle{headings}
\mainmatter
\def\ECCVSubNumber{2688}  

\title{Toward Understanding WordArt: Corner-Guided Transformer for Scene Text Recognition} 

\titlerunning{Toward Understanding WordArt: Corner-Guided Transformer for STR}

\author{Xudong Xie\inst{1} \and
Ling Fu\inst{1} \and
Zhifei Zhang\inst{2} \and
Zhaowen Wang\inst{2} \and
Xiang Bai\inst{1}\textsuperscript{(\Letter)}}
\authorrunning{X. Xie et al.}
%
\institute{Huazhong University of Science and Technology, China
\email{\{xdxie,ling\_fu,xbai\}@hust.edu.cn} \and
Adobe Research, USA \\
\email{\{zzhang,zhawang\}@adobe.com}}
\maketitle

\def\thefootnote{\Letter}\footnotetext{Corresponding author}\def\thefootnote{\arabic{footnote}}

\begin{abstract}

Artistic text recognition is an extremely challenging task with a wide range of applications. However, current scene text recognition methods mainly focus on irregular text while have not explored artistic text specifically. The challenges of artistic text recognition include the various appearance with special-designed fonts and effects, the complex connections and overlaps between characters, and the severe interference from background patterns. 
To alleviate these problems, we propose to recognize the artistic text at three levels. Firstly, corner points are applied to guide the extraction of local features inside characters, considering the robustness of corner structures to appearance and shape. In this way, the discreteness of the corner points cuts off the connection between characters, and the sparsity of them improves the robustness for background interference. 
Secondly, we design a character contrastive loss to model the character-level feature, improving the feature representation for character classification.
Thirdly, we utilize Transformer to learn the global feature on image-level and model the global relationship of the corner points, with the assistance of a corner-query cross-attention mechanism.
Besides, we provide an artistic text dataset to benchmark the performance. 
Experimental results verify the significant superiority of our proposed method on artistic text recognition and also achieve state-of-the-art performance on several blurred and perspective datasets. The dataset and codes are available at: \href{https://github.com/xdxie/WordArt}{https://github.com/xdxie/WordArt}.


\end{abstract}

\section{Introduction}

\begin{figure}
\centering
\includegraphics[width=0.81\textwidth]{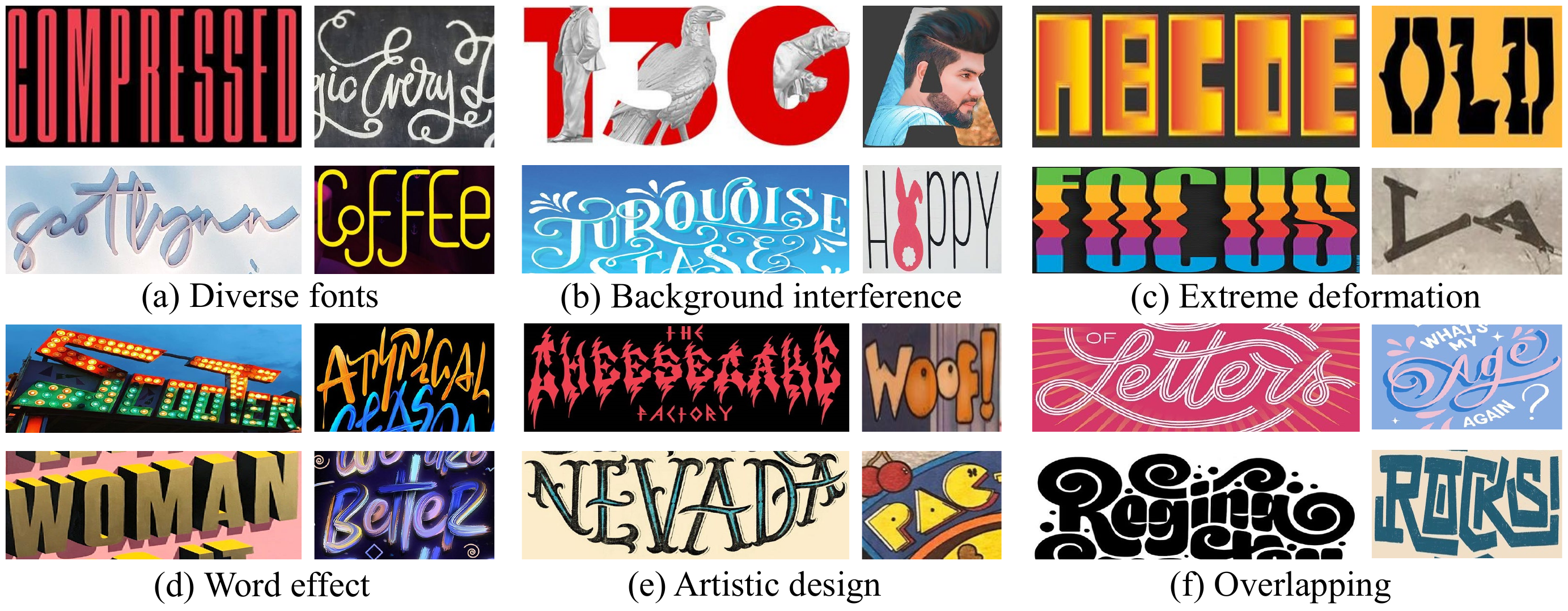}
\caption{The artistic text examples of different types from the WordArt dataset }
\label{fig:example}
\end{figure}

The artistic text is a kind of beautified text that is carefully designed by designers or artists. They use various complex fonts of different styles, combining word effects such as shadow, rotation, stereo transformation, deformation, and distortion. Meanwhile, the background patterns and text meaning are considered during the design. Artistic text is widely used in advertisements, slogans, exhibitions, decorations, magazines, and books. Fig.~\ref{fig:example} shows some typical artistic text images with several unique properties.

In view of this, artistic text recognition is an overlooked and extremely challenging task with importance and practicability in a wide range of applications. Unlike scene text recognition (STR)~\cite{shi2016end,shi2018aster,fang2021read}, artistic text recognition often has several difficulties and challenges: (1) As illustrated in Fig.~\ref{fig:example} (a, c, d), the character appearance varies widely due to the different fonts, artistic design effects, and deformation. (2) In Fig.~\ref{fig:example} (a, f), there are many complicated connections and overlaps between characters, which makes it difficult to focus on the center or the stroke of a character independently during the recognition process. (3) The design of the artistic text may use background elements to express characters or words and organically combine texts with patterns, causing serious background interference, as shown in Fig.~\ref{fig:example} (b, e).

It is difficult for existing scene text recognition methods to be competent for this task. The approaches for regular scene text~\cite{graves2006connectionist,shi2016end,he2016reading,cheng2017focusing} only focus on horizontal text with standard printing fonts and cannot cope with instances with various shapes, artistic effects, and fonts. Other methods utilizing rectification~\cite{shi2016robust,shi2018aster,luo2019moran} for irregular scene text recognition can rectify the text line but not the various character shapes. The existing methods based on the attention mechanism~\cite{yang2017learning,li2019show,lee2020recognizing} cannot obtain accurate positions of artistic characters, as shown in Fig.~\ref{fig:feature}. 
In a sense, irregular texts belong to a subset of artistic texts. In addition, handwriting contains a variety of fonts and ligatures, but the background of these instances is very simple without word effects and artistic designs. Therefore, the methods for handwriting recognition~\cite{frinken2015deep,bhunia2021metahtr} fail to handle the artistic text with complex background. Recently, some researchers have introduced linguistic knowledge and corpora to help improve the performance of scene text recognition~\cite{qiao2020seed,fang2021read}. However, as shown in Fig.~\ref{fig:qualitative}, the language model is also inefficient for the complex artistic text. Therefore, we need to learn more robust and representative visual features.

Considering these challenges, in order to obtain robust visual features to recognize the artistic text accurately, we propose to model image features at three levels. 
{\bf (1) Local feature within character.} In the artistic text, the appearance and shape of characters vary widely from instance to instance. It is necessary to build an explicit invariant feature within characters to robustly represent the core key points or structures, suppressing the interference of appearance and deformation. Since the corner points~\cite{shi1994good,harris1988combined} of the character strokes and the relative positions between these points are invariant, we use the corner point map as a robust representation of the input image. Moreover, the discreteness of the corner point map cuts off the connection and the overlap between characters, and the sparsity suppresses most of the background interference. In addition, we propose a corner-query cross-attention mechanism, treating the corner point as the query and the image as the key to make the corner seek the image features of interest. In this way, the corner guides the model to pay more accurate attention to the core strokes or character centers of the artistic text. 
{\bf (2) Character-level feature.} Accurate character recognition
is critical for text recognition. The huge visual differences between the same characters of artistic texts lead to the scattered distribution of their features in the feature space. To implicitly learn common representations for each class of characters, it is necessary to make the same class instances cluster together in the feature space and different classes away from each other. Therefore, we introduce a loss function based on the contrastive learning~\cite{chen2020simple,khosla2020supervised}, significantly improving the clustering degree of their features (Fig.~\ref{fig:loss}). For each character in a minibatch, its positive samples are characters of the same class, and other characters are negative samples.
{\bf (3) Global feature on image-level.} Global features of images can assist the overall text recognition because the characters can be reasoned through the visual and semantic information from context. Transformer~\cite{vaswani2017attention} based on the self-attention mechanism has demonstrated its strong advantages and performance~\cite{lee2020recognizing,fang2021read}, benefiting from its global modeling ability. Therefore, to extract the global features of artistic text images with arbitrary shapes and model the global relationship of the corner points, we use Transformer~\cite{vaswani2017attention} as our backbone and propose the CornerTransformer.

To benchmark the performance of different methods on the artistic text recognition task, we propose a dataset named WordArt. Experimental results show that our method outperforms the existing STR models on this challenging task. CornerTransformer performs well on many artistic texts containing complex fonts, ligatures, and overlaps. Furthermore, we achieve competitive or better results than other methods on common STR benchmarks. In particular, our model outperforms the SOTAs on Street View Text~\cite{wang2011end}, SVT-Perspective~\cite{phan2013recognizing} and ICDAR 2015~\cite{karatzas2015icdar} benefits from the corner point map, as gradient-based corner point detection is robust to image resolution, noise and blur.

To summarize, the contributions of this paper are four-fold:
\begin{enumerate}
\item[(1)] We focus on a new challenging task: artistic text recognition, and propose the WordArt dataset to benchmark the performance.    
\item[(2)] We notice the importance of the corner point on artistic text recognition and present a corner-query cross-attention mechanism, which allows the model to pay more accurate attention to the core strokes or character centers.
\item[(3)] We design a character contrastive loss to cluster the same class of character features, enabling the model to learn unified representations for characters.
\item[(4)] Our method significantly outperforms other models on artistic text recognition and also achieves new state-of-the-art results on scene text recognition.
\end{enumerate}

\section{Related Work}
\textbf{Scene text recognition.}
Scene text can be roughly divided into regular and irregular text. The sequence-to-sequence models based on CTC~\cite{graves2006connectionist,shi2016end,he2016reading} and attention~\cite{cheng2017focusing} for regular text recognition have made a great progress. However, these methods fail to cope with curved or rotated text, so irregular text recognition has recently attracted many research interests. The rectification-based methods~\cite{shi2016robust,shi2018aster,luo2019moran,yang2019symmetry} utilize the spatial transformer network~\cite{jaderberg2015spatial} to transform the text image into a canonical shape, but the predefined transformation space limits the generalization of them. The segmentation-based methods ~\cite{liao2019scene,wan2020textscanner} formulate the
recognition task as a character segmentation problem, but character-level annotations are required. In addition, the recent methods with the 2D attention mechanism~\cite{yang2017learning,li2019show,lee2020recognizing} also show strong performance on irregular text recognition, and we choose SATRN~\cite{lee2020recognizing} as the baseline to build our model. Overall, it is difficult to directly apply these methods to artistic text recognition because of the limitations stated in Sec. 1.

\noindent\textbf{Special text recognition.}
Beyond scene text recognition, other recognition tasks for special text are also significantly important. For example, handwriting recognition~\cite{hu1996hmm,zhang2017drawing,bhunia2021metahtr} has always been the focus of research given the changeable character shapes and varying writing styles.
Another meaningful task that has emerged recently is handwritten mathematical expression recognition~\cite{wu2020handwritten,le2020recognizing}, which has wide applications in education. Manga text recognition~\cite{gobbo2020unconstrained} is also an interesting problem due to the unconstrained text in the manga. Moreover, Wang~\emph{et al.}~\cite{wang2020exploring} specifically explore font-independent features of scene texts via a glyph generative adversarial network~\cite{goodfellow2014generative}. Compared with the artistic text, the backgrounds of handwriting and manga images are very simple, and there is no character overlapping, artistic rendering, or word effects. To our knowledge, ours is the first work for artistic text recognition.

\noindent\textbf{Text recognition with auxiliary information.} 
Some segmentation-based methods ~\cite{liao2019scene,wan2020textscanner} introduce character-level annotations to improve the recognition results. Other recent approaches~\cite{qiao2020seed,fang2021read} transfer linguistic knowledge to the vision model with a pre-trained language model. Through linguistic information, the model can predict characters according to the context. However, to utilize such information needs to pay the extra cost of data and computing. Besides the deep learning-based methods, other traditional methods explore robust text image presentations, such as SIFT descriptors~\cite{phan2013recognizing}, Strokelets~\cite{yao2014strokelets}, and HOG~\cite{su2014accurate}. Access to such information is automatic and almost cost-free. In this paper, we use the corner point~\cite{shi1994good,harris1988combined} to assist the Transformer-based~\cite{vaswani2017attention,lee2020recognizing} method for artistic text recognition.

\noindent\textbf{Text recognition dataset.} 
There exist several standard datasets for the task of scene text recognition. IIIT5k-Words (IIIT5k)~\cite{mishra2012scene}, ICDAR 2013 (IC13)~\cite{karatzas2013icdar}, and Street View Text (SVT)~\cite{wang2011end} only contain horizontal text with standard fonts. ICDAR 2015 (IC15)~\cite{karatzas2015icdar} contains many small, blurred, and irregular text. SVT-Perspective (SVTP)~\cite{phan2013recognizing} is built based on the original SVT to evaluate perspective distorted text recognition. CUTE80 (CUTE)~\cite{risnumawan2014robust} and Total-Text~\cite{ch2020total} mainly focus on curved text. COCO-Text~\cite{veit2016coco} is the first large-scale dataset for text in natural images. Besides, there are some multilingual datasets such as CTW~\cite{yuan2019large}, LSVT~\cite{sun2019icdar} and MLT~\cite{nayef2019icdar2019}. However, most images in these datasets do not contain artistic text. Therefore, we construct a new dataset to benchmark the performance of artistic text recognition.

\section{Methodology}


\subsection{Overview}
\begin{figure}
\centering
\includegraphics[width=1\textwidth]{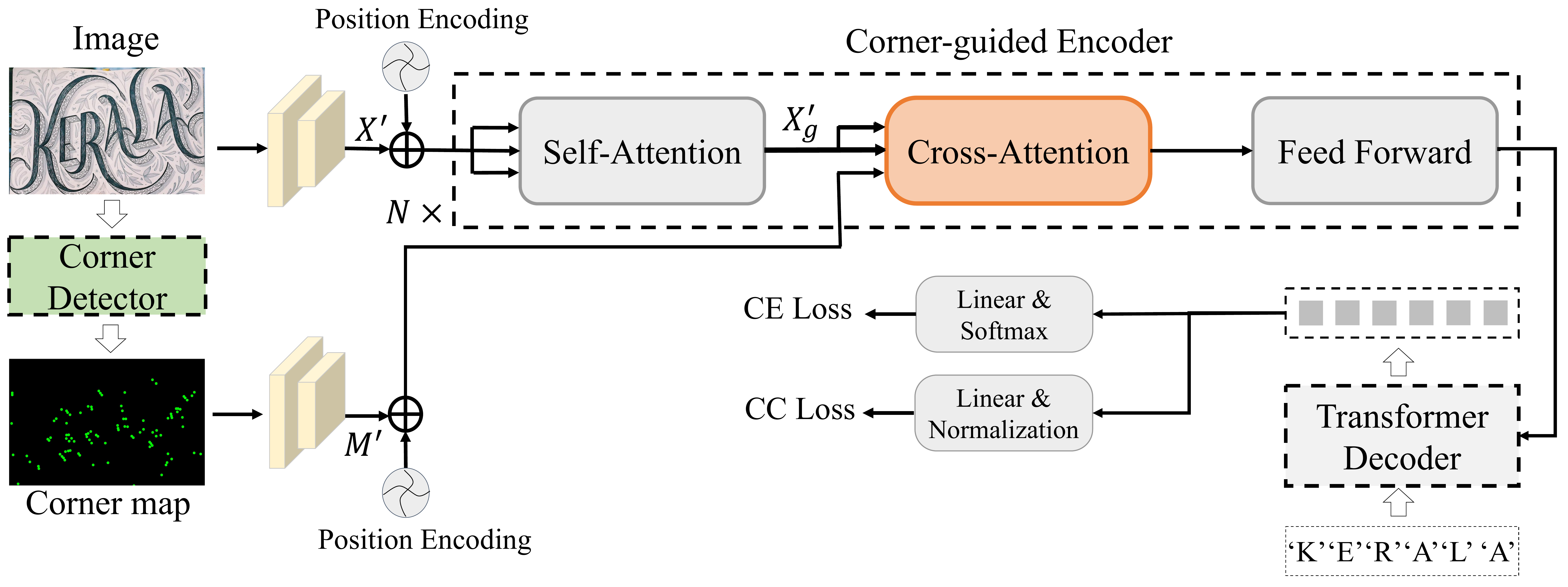}
\caption{The overall architecture of CornerTransformer consists of two inputs from different modalities, a corner-guided encoder and a Transformer decoder~\cite{vaswani2017attention}. CE loss is the cross-entropy loss, and CC loss is our proposed character contrastive loss}
\label{fig:pipeline}
\end{figure}

The overall structure of CornerTransformer is shown in Fig.~\ref{fig:pipeline}. Given an image $X \in \mathbb{R}^{H\times W\times 3}$, we first utilize a corner point detector to generate a corner point map $M \in \mathbb{R}^{H\times W\times 1}$. Then, $X$ and $M$ are fed into two convolutional layers respectively for producing features of $X^{'} \in \mathbb{R}^{{\frac{H}{4}}\times {\frac{W}{4}}\times C}$ and $M^{'} \in \mathbb{R}^{{\frac{H}{4}}\times {\frac{W}{4}}\times C}$, where $C$ is the feature dimension. On the one hand, $X^{'}$ will learn the global features $X^{'}_g$ of the image through the multi-head self-attention mechanism. On the other hand, $M^{'}$ will combine with $X^{'}_g$ through the multi-head cross-attention mechanism.
Then, the encoder output feature and the character sequence embedding will be fed into the Transformer decoder~\cite{vaswani2017attention} to generate the feature sequence. Finally, we apply two linear branches to calculate the cross-entropy loss and the character contrastive loss separately.

\subsection{Corner-guided Encoder}

\begin{figure}
\centering
\includegraphics[width=0.9\textwidth]{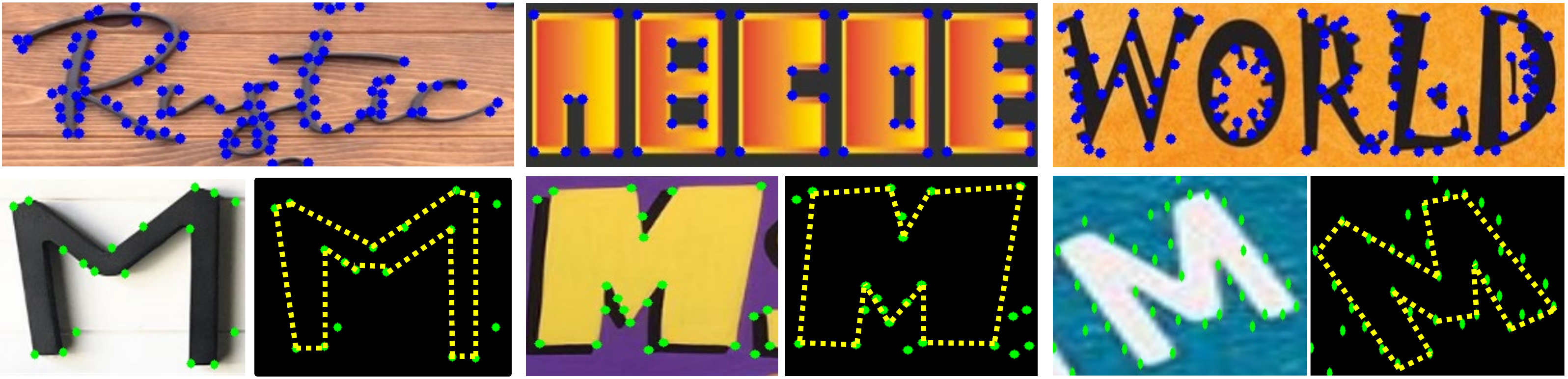}
\caption{Visualization of corner point detection. \textbf{Top:} The detected corner points of artistic text images.
\textbf{Bottom:} Corner points of a character “M” with various appearances, whose structural relations are similar}
\label{fig:corner}
\end{figure}

In the task of artistic text recognition, the deformation and distortion of characters are extremely diverse due to various fonts and artistic effects. Thus, it is necessary to transform the artistic text image into a more robust representation. As shown in Fig.~\ref{fig:corner}, we observe that, for the great variance in the appearance of a specific character, the most critical corners of this character can almost always be detected. The structural relations formed by the connection of these key corners are similar. Moreover, these points are the positions that contain rich visual information of the image.
Therefore, we utilize the corner map extracted from the image as an auxiliary input to provide an invariant visual representation. In addition, the connection and the position overlap between characters are extremely complicated, while the discrete corner map can naturally cut off the connection and suppress the overlapping effect of strokes. Furthermore, designers often use some background elements when designing the artistic text to perfectly integrate the artistic text with the background, which causes serious interference from the background during the recognition process. However, the corner map only retains the keypoints of the image, suppressing most background elements and making it easier for the model to focus on important text features.

Given an image, we use a classical corner point detector, Shi-Tomasi corner detector~\cite{shi1994good}, to generate its corner map. This detector improves the stability of Harris detector~\cite{harris1988combined}, and can produce high-quality corner points. For each pixel $(x, y)$ in the image, we first calculate the image structure tensor $S$, then the corner response function is defined as $R = \min(\lambda_1, \lambda_2)$, where $\lambda_1$ and $\lambda_2$ are the eigenvalues of $S$. If $R > threshold$, pixel $(x, y)$ is a corner point and the value in position $(x, y)$ of the corner map is $1$, otherwise it is $0$. Therefore, the corner map is a sparse matrix whose element of value $1$ only represents the position information of corners.

After obtaining the corner map, considering that there are local correlations between corners instead of being independent of each other, we first use two convolutional layers to model local relations on the corner map and add 2D position encoding~\cite{lee2020recognizing} to record the corner position information. A natural method to combine image and corner features is to concatenate them together and feed them into the Transformer encoder. However, this can not make full use of the auxiliary information of corners as shown in Tab.~\ref{table:fusion}. Since the corner map is sparse, the model will still mainly focus on image features. Therefore, we design a corner-guided encoder to fuse corner features at each block. Specifically, we add a multi-head cross-attention layer after the self-attention layer. We utilize the image feature $X^{'}_g$ as the key and value, and the corner feature $M^{'}$ as the query. The corner-query cross-attention mechanism can be formulated as:
\begin{equation}
CA(Q, K, V) = CA(M^{'}, X^{'}_g, X^{'}_g) = softmax(\frac{M^{'}{X^{'}_g}^T}{\sigma})X^{'}_g,
\end{equation}
where $CA$ means Cross-Attention and $\sigma$ is a scaling factor. Since corners represent keypoints inside characters, we use the corner map as a query to make the corner seek the image features of interest. Furthermore, the model can pay more accurate attention to the character positions of the artistic text in the image. For instance, for character ``A'' in a text image, its top corner point tends to focus on other positions of this character rather than other characters. Our ablation study and visualization analysis also prove the effectiveness of the corner-query cross-attention mechanism. 

The corner-guided encoder is composed of a stack of $N$ blocks, where each consists of a self-attention layer, a cross-attention layer, and a feed-forward layer. The query of each cross-attention layer is $M^{'}$.

\subsection{Character Contrastive loss}
Corner-based representation mainly focuses on the local modeling within the character, while Transformer tends to the global modeling of the whole image. To bridge these two representation levels, we introduce a middle-level (character-level) representation learning method. For the artistic text, different instances of the same character show a variety of appearances, including font, shadow, rotation, and other effects. Therefore, in the training process, it is necessary to learn an implicit and unified character-level representation for each character class, so that instances of the same character class are clustered together in the feature space, and features of different classes are far away from each other. 

Inspired by the popular thought of contrastive learning~\cite{chen2020simple,khosla2020supervised,zhang2022context}, we propose a Character Contrastive loss (CC loss) to achieve our motivation. In short, for a character in a minibatch, the characters of the same class are positive samples, and other characters are negative samples. Specifically, given a minibatch of $N$ images, each image contains variable-length text. We unify the length of text labels to $m=25$, and there are $N\times m$ characters in a minibatch. For the $i$th character, $x_i$ is the feature vector and $y_i$ is the class label, where $i \in I \equiv \{1, 2, ..., N\times m\}$. When the $i$th character is an anchor, its positive set is $P(i) \equiv \{p\in I:y_p = y_i, p \neq i\}$, and the negative set is $N(i) \equiv \{n\in I:y_n \neq y_i, n \neq i\}$. The character contrastive loss can be formulated as:
\begin{equation}
\label{loss}
\mathcal{L}_{\mathrm{CC}} = \sum_{i \in I}{\frac{-1}{N_p}}\sum_{p \in P(i)}log\frac{exp(x_i\cdot x_p/\tau)}{\sum\nolimits_{s\in P(i)}exp(x_i\cdot x_s/\tau) + \sum\nolimits_{t\in N(i)}exp(x_i\cdot x_t/\tau)},
\end{equation}
where $N_p$ is the number of positive samples, and $\tau$ a scaling factor. 

Finally, the full optimization objective is defined as:
\begin{equation}
	\mathcal{L}= \mathcal{L}_{\mathrm{CE}} + \lambda \mathcal{L}_{\mathrm{CC}},
\end{equation}
where $\mathcal{L}_{\mathrm{CE}}$ is the cross entropy loss. We set $\lambda = 0.1$ by default.

\section{Experiments}

\subsection{WordArt Dataset}
To benchmark the performance of different models on the artistic text recognition task, we collect a dataset of artistic text named WordArt. Thanks to the TextSeg~\cite{xu2021rethinking} dataset, which contains images of posters, greeting cards, covers, billboards, handwriting, etc. There exist many artistic texts in these images. In view of this, we first crop the word images with the word bounding box annotations and then carefully pick over the artistic text following the definition of the artistic text as stated in Sec. 1. Finally, our WordArt dataset consists of 6316 artistic text images. Following the splitting rule of TextSeg, the training set contains 4805 images, and the testing set contains 1511 images. The statistical analysis is presented in Fig.~\ref{fig:dataset}. The distributions of text length and character frequency roughly align with the English corpus.
The qualitative presentation of the WordArt dataset is shown in Fig.~\ref{fig:example}. 

{
\begin{figure}

\centering
\subfigure[Text length distribution]{\includegraphics[width = 0.4\textwidth]{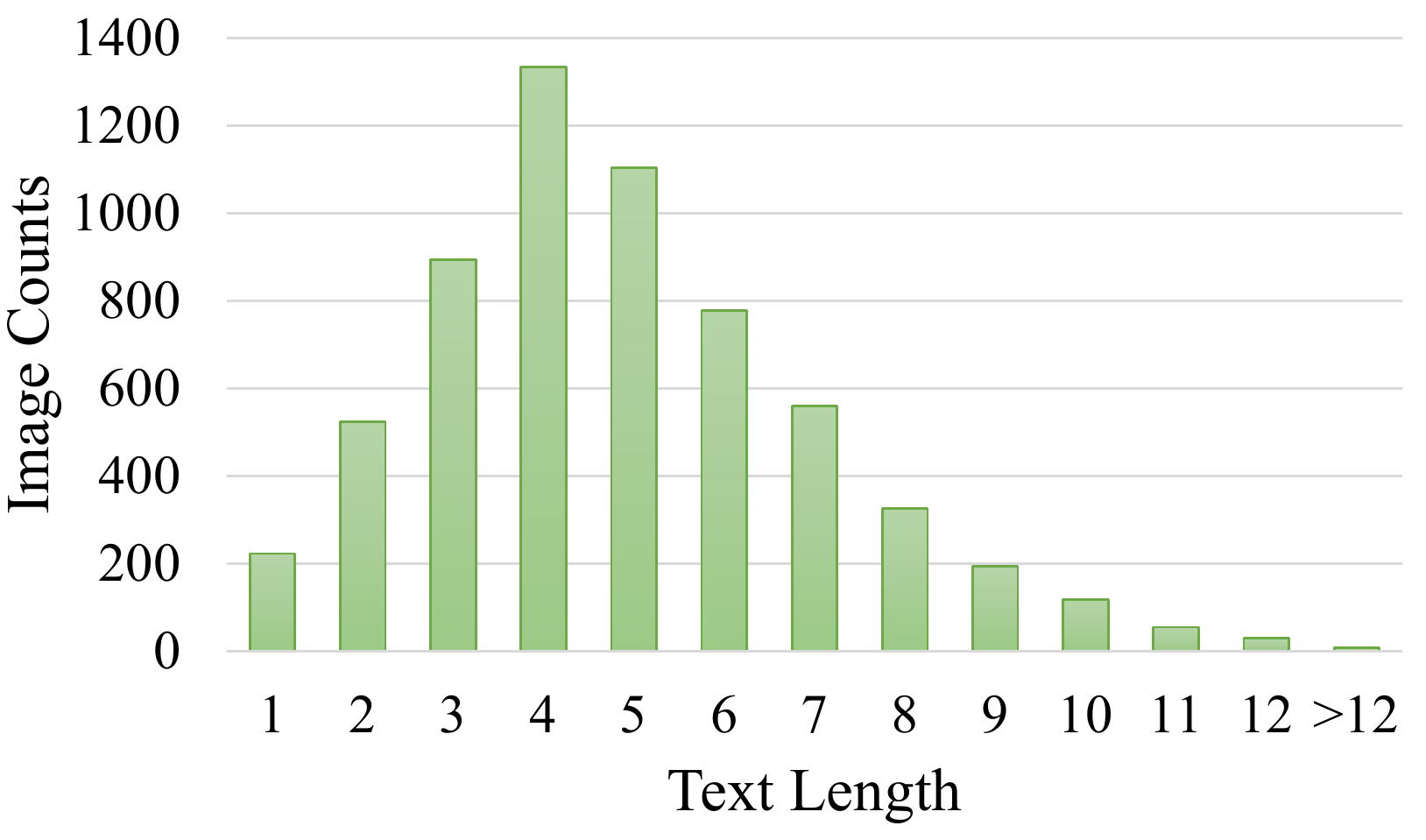}}
    \label{fig:short-a}
\subfigure[Character frequency]{\includegraphics[width = 0.55\textwidth]{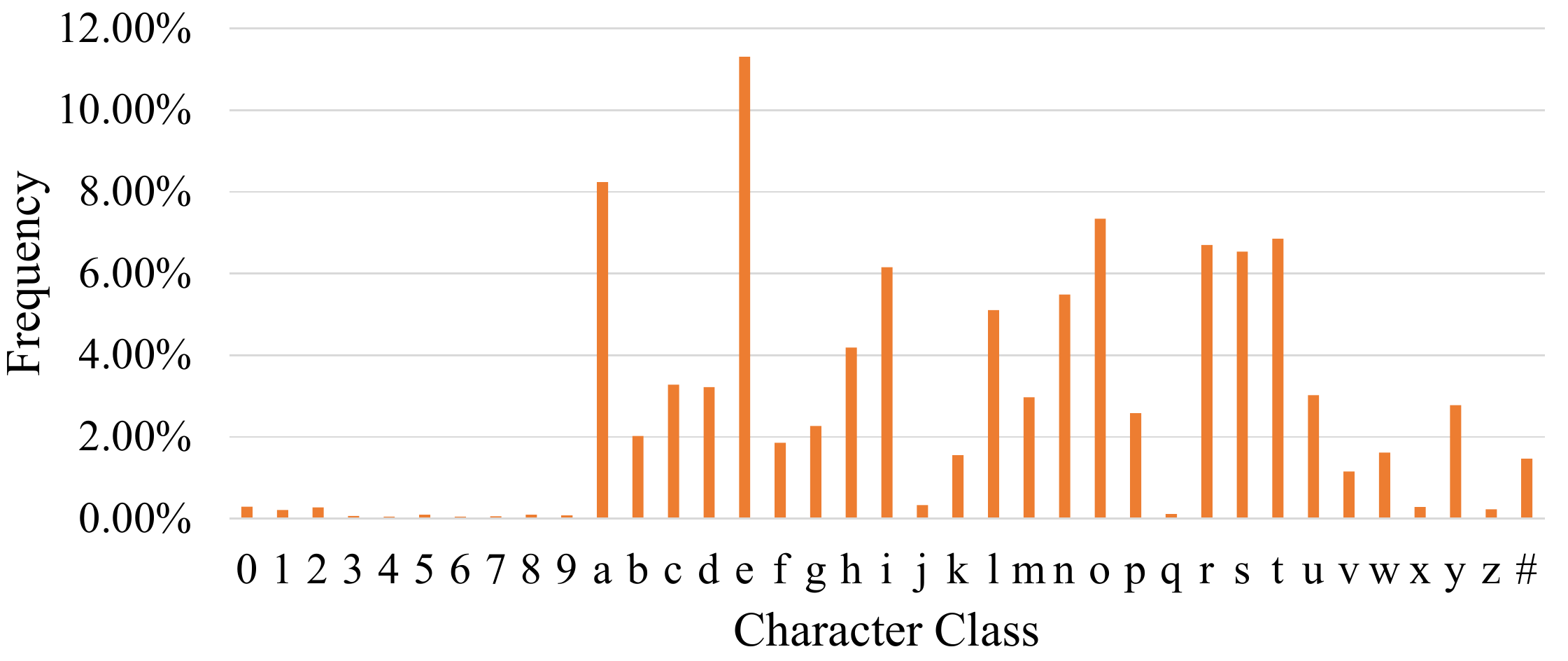}}
    \label{fig:short-b}
\caption{Statistical analysis for the WordArt dataset. (a) The number of images with different text lengths. (b) Frequency distribution of all characters in the whole dataset} 
\label{fig:dataset}
\end{figure}
}

\subsection{Implementation Details}
In our CornerTransformer, the feature dimension of all the attention layers is set to 512, with 8 heads for each layer. We set $N=12$ for the corner-guided encoder. To calculate the character contrastive loss, we add two linear layers with 2048 hidden dimension and an $L2$ normalization to transform the decoder output features into a normalized feature space. By default, we jointly use CE loss and CC loss to train our model. $\tau$ in CC loss is set to 0.1. As common practice~\cite{yu2020towards,baek2019wrong}, we use two synthetic datasets MJSynth (MJ)~\cite{2014Synthetic,jaderberg2016reading} and SynthText~\cite{gupta2016synthetic} as training datasets and directly evaluate the performance on the WordArt dataset and STR datasets after training. The input images are resized to $32\times 128$ for both training and testing with batch size 256. The model is trained with Adam optimizer~\cite{kingma2014adam} with the initial learning rate $3e^{-4}$. The total epoch is 6, and the learning rate will decay to $3e^{-5}$ after 4 epochs. We adopt several data augmentation strategies such as rotation, Gaussian noise, etc.

\subsection{Ablation study}
For artistic text recognition, our main contributions to the method are introducing the corner map with designing the corner-query cross-attention and proposing the character contrastive loss. We will verify the effectiveness of each design in detail. Since we need to use the global modeling capabilities of Transformer~\cite{vaswani2017attention}, we choose SATRN~\cite{lee2020recognizing} as our baseline and reproduce its model by replacing the dimensions the query and key from 128 to 512. To comprehensively evaluate different designs, besides the word accuracy, we present character recall and character precision to assist the evaluation. All the results for the ablation study are evaluated on our WordArt dataset.

\noindent\textbf{The effectiveness of the corner map.} 
Since we add an attention module to fuse the corner map, extra parameters are introduced, increasing the capacity of the model. In order to verify that the performance improvement comes from the role of the corner map rather than the extra parameters, we replace the input of the corner branch in Fig.~\ref{fig:pipeline} with the same image as the main branch. The results shown in the third row of Tab.~\ref{table:corner} are lower than baseline. We attribute this to the large amount of noise and redundant information contained in the image, which adds false guidance to the model when directly applying it as a query. Besides, we also remove the extra input branch but retain the added attention module. In this case, the cross-attention turns into self-attention. As shown in the fourth row of Tab.~\ref{table:corner}, this only gives a slight boost to the baseline results. Therefore, the role of the corner map is crucial for artistic text recognition, indicating keypoints and filtering out noise.

\setlength{\tabcolsep}{4pt}
\begin{table}
\begin{center}
\caption{Ablation study about the effectiveness of the corner map}
\label{table:corner}
\scalebox{0.9}{
\begin{tabular}{cccc}
\hline\noalign{\smallskip}
Input & word acc & char recall & char precision \\
\noalign{\smallskip}
\hline
\noalign{\smallskip}
Baseline (Self-attn) & 67.0 & 84.6 & 84.2 \\
Corner+Image & {\bf 69.1}  & {\bf 85.7} & {\bf 84.8} \\
Image+Image & 66.0 & 83.8 & 83.3\\
Self-attn $\times 2$ & 67.6  & 85.2 & 83.3\\
\hline
\end{tabular}
}
\end{center}
\end{table}
\setlength{\tabcolsep}{1.4pt}

\noindent\textbf{Different corner detectors.}
In view of the importance of the corner map for model performance, 
it is necessary to choose a suitable corner detector to obtain high-quality corner maps. The detector used in our model is the Shi-Tomasi corner detector~\cite{shi1994good}. We also experiment with the Harris detector~\cite{harris1988combined} but found it often produces more extra noise corners, which has slight damage to performance. In addition, we use a deep learning-based corner detector SuperPoint~\cite{detone2018superpoint}. We load its pre-trained model to produce corner maps, and the results are presented in Tab.~\ref{table:detector}. Although SuperPoint can generate high-quality corner maps, it uses an additional neural network model that increases the feed-forward time. 
\setlength{\tabcolsep}{4pt}
\begin{table}
\begin{center}
\caption{Results of different corner detectors}
\label{table:detector}
\scalebox{0.9}{
\begin{tabular}{cccc}
\hline\noalign{\smallskip}
Corner Detector & word acc & char recall & char precision \\
\noalign{\smallskip}
\hline
\noalign{\smallskip}
Shi-Tomasi~\cite{shi1994good} & {\bf 69.1}  & {\bf 85.7} & {\bf 84.8} \\
Harris~\cite{harris1988combined} & 68.4  & 85.1 & 84.6\\
SuperPoint~\cite{detone2018superpoint} & 69.0 & 85.3 & 84.7\\
\hline
\end{tabular}
}
\end{center}
\end{table}
\setlength{\tabcolsep}{1.4pt}
\setlength{\tabcolsep}{4pt}
\begin{table}
\begin{center}
\caption{Results of different fusion strategies}
\label{table:fusion}
\scalebox{0.9}{
\begin{tabular}{cccc}
\hline\noalign{\smallskip}
Fusion strategy & word acc & char recall & char precision \\
\noalign{\smallskip}
\hline
\noalign{\smallskip}
Baseline & 67.0 & 84.6 & 84.2 \\
Corner-query & {\bf 69.1}  & {\bf 85.7} & {\bf 84.8} \\
Corner-key/value & 66.9 & 84.1 & 84.2\\
Concat & 67.0  & 84.7 & 84.4\\
Add & 66.6 & 84.9 & 84.3\\
Multiply & 67.4 & 84.7 & 84.4\\
\hline
\end{tabular}
}
\end{center}
\end{table}
\setlength{\tabcolsep}{1.4pt}
It is worth noting that no matter which detector we use, they can all capture the most critical corner locations and the structure of the text. Therefore, the results in Tab.~\ref{table:detector} using corner maps are better than the other results in Tab.~\ref{table:corner}.
\noindent\textbf{Fusion strategy.}
It is crucial to efficiently fuse the features of the corner map and the image, which determines whether the model can make full use of the important information carried by the corners. Given these two features obtained from convolutional layers, we can fuse them into one by Concat, Add and Multiply operations and straightly feed the fused feature to Transformer. As shown in Tab.~\ref{table:fusion}, there is no significant improvement in these results. Add operation introduces additive noise to image features. Multiply operation makes the image filter out valuable features based on the corners, bringing a slight improvement. Moreover, for the cross-attention module, we swap the roles of corner and image, so that corner features are used as the key and value. But the results are not improved, although this operation introduces extra parameters compared to the baseline. The reason is that a lot of information is lost when the corner map is used as the value. Therefore, our corner-query cross-attention mechanism is an efficient fusion strategy.

\setlength{\tabcolsep}{4pt}
\begin{table}
\begin{center}
\caption{Ablation study on character contrastive loss}
\label{table:loss}
\scalebox{0.9}{
\begin{tabular}{cccc}
\hline\noalign{\smallskip}
Hyperparameters & word acc & char recall & char precision \\
\noalign{\smallskip}
\hline
\noalign{\smallskip}
$\lambda=0$ (without CC loss) & 67.0 & 84.6 & 84.2 \\
$\lambda=0.1$, $\tau=0.05$, $d=512$ & 66.5 & 84.0 & 83.6 \\
$\lambda=0.1$, $\tau=0.1$, $d=512$ & 68.1 & 85.5 & 85.3\\
$\lambda=0.1$, $\tau=0.15$, $d=512$ & 67.7 & 84.9 & 84.6\\
$\lambda=0.1$, $\tau=0.1$, $d=2048$ & {\bf 68.6} & {\bf 85.8} & {\bf 85.9}\\
$\lambda=0.01$, $\tau=0.1$, $d=2048$ & 67.2 & 84.4 & 84.3\\
$\lambda=1$, $\tau=0.1$, $d=2048$ & 66.6 & 83.9 & 83.7\\
\hline
\end{tabular}
}
\end{center}
\end{table}
\setlength{\tabcolsep}{1.4pt}

\noindent\textbf{Character contrastive loss.}
According to the previous work of contrastive learning~\cite{chen2020simple,khosla2020supervised}, the scaling factor $\tau$ of the loss function in formula~(\ref{loss}) plays an important role in final performance. Relatively low values of $\tau$ make hard negatives have more weight but the feature space will be less smooth when $\tau$ is extremely low. We conduct an ablation study on $\tau$ as shown in Tab.~\ref{table:loss}, and found $\tau=0.1$ is optimal. Besides, the dimension of the final output feature vector $x_i$ also affects performance. Generally, higher dimension brings better results because the feature vector represents more information. If the weight of the CC loss is small ($\lambda=0.01$), it will not bring a significant performance improvement. In contrast, if $\lambda=1$, it will interfere the joint optimization, resulting in performance degradation.
As a result, we adopt $\lambda=0.1$, $\tau=0.1$, $d=2048$ in our model. The results of character recall and character precision show that CC loss actually improves the performance of character recognition.

\subsection{Performance for Artistic Text Recognition }

\begin{figure}
\centering
\includegraphics[width=0.9\textwidth]{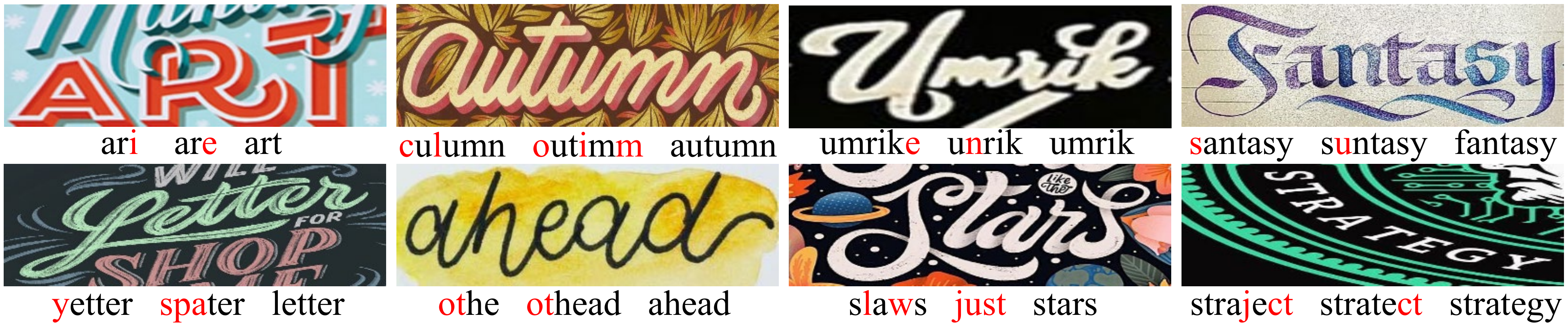}
\caption{Qualitative recognition results on WordArt dataset. Each example is along with the results from ABINet-LV~\cite{fang2021read}, our baseline and the proposed CornerTransformer, separately. Hard examples successfully recognized by CornerTransformer}
\label{fig:qualitative}
\end{figure}

In order to demonstrate the superiority of our CornerTransformer on the artistic text recognition task, we compare it with several state-of-the-art scene text recognition methods in Tab.~\ref{table:wordart_res}. All the results of these methods are obtained by directly loading their released checkpoints to be evaluated on WordArt. Our CornerTransformer shows a significant superiority, thanks to the corner-query cross-attention and the character contrastive loss. Fig.~\ref{fig:qualitative} presents some hard examples that are successfully recognized by CornerTransformer. Our model can cope with artistic texts containing complex fonts, ligatures, overlaps, and many extremely curved and deformed texts.

\setlength{\tabcolsep}{4pt}
\begin{table}
\begin{center}
\caption{Performance comparison with other methods on WordArt dataset. * indicates the baseline of SATRN\cite{lee2020recognizing} reimplemented by ourselves, replacing the dimensions of the query and key from 128 to 512. Inference time is estimated using an NVIDIA TITAN Xp by averaging 3 trials, based on Pytorch implementation. $^\dag$ indicates the inference time is estimated based on the TensorFlow implementation. ``WiKi'' indicates using a language model trained with WiKiText-103~\cite{merity2016pointer}.}
\label{table:wordart_res}
\scalebox{0.9}{
\begin{tabular}{ccccc}
\hline\noalign{\smallskip}
Methods & Training Data & WordArt & Params (M) & Time (ms) \\
\noalign{\smallskip}
\hline
\noalign{\smallskip}
CRNN~\cite{shi2016end} & ST+MJ  & 47.5 & 8.3 & 9.9 \\
ASTER~\cite{shi2018aster} & ST+MJ  & 57.9 & 21 & 247.9 \\
TRBA~\cite{baek2019wrong} & ST+MJ  & 55.8 & 49.6 & 28.8 \\
DAN~\cite{wang2020decoupled} & ST+MJ  & 52.4 & 18.2 & 41.7\\
NRTR~\cite{sheng2019nrtr} & ST+MJ  & 58.5 & 66.7 & 350.8\\
RobustScanner~\cite{yue2020robustscanner} & ST+MJ+SA+R & 61.3 & 48.0 & 71.0\\
SAR~\cite{li2019show} & ST+MJ+SA+R  & 63.8 & 57.5 & 109.2\\
SEED~\cite{qiao2020seed} & ST+MJ & 60.1 & 25.0 & 158.8\\
SCATTER~\cite{litman2020scatter} & ST+MJ+SA  & 64.0 & 119.7 & 142.7\\
SATRN$^\dag$~\cite{lee2020recognizing} & ST+MJ  & 65.7 & 55.5 & 494.1\\
ABINet-LV~\cite{fang2021read} & ST+MJ+WiKi  & 67.4 & 36.7 & 42.4\\
\noalign{\smallskip}
\hline
\noalign{\smallskip}
Baseline* & ST+MJ  & 67.0 & 65.6 & 274.7\\
Baseline + Corner & ST+MJ & 69.1 & 80.5 & 294.9\\
Baseline + CC loss & ST+MJ & 68.6 & 70.9 & 274.7\\
CornerTransformer & ST+MJ & {\bf 70.8} & 85.7 & 294.9\\
\hline
\end{tabular}
}
\end{center}
\end{table}
\setlength{\tabcolsep}{1.4pt}

\subsection{Evaluation on STR Benchmarks}
To further verify the generalization of CornerTransformer, we also conduct evaluations on six STR benchmarks: IIIT5k~\cite{mishra2012scene}, IC13~\cite{karatzas2013icdar}, SVT~\cite{wang2011end}, IC15~\cite{karatzas2015icdar}, SVTP~\cite{phan2013recognizing} and CUTE~\cite{risnumawan2014robust}.
The results compared with other state-of-the-art methods are shown in Tab.~\ref{table:str_res}. We can achieve state-of-the-art results on SVT and IC15 because most images are severely corrupted by noise and blur, while gradient-based corner detection is robust to image resolution, noise and blur. Besides, we also obtain a competitive result on CUTE and the best result on SVTP. The texts in these datasets are perspective and curved, while the relative position between corner points is invariant.

\setlength{\tabcolsep}{2pt}
\begin{table}
\begin{center}
\caption{Accuracy comparison with other STR methods on six standard benchmarks} 
\label{table:str_res}
\scalebox{0.9}{
\begin{tabular}{cccccccccc}
\hline\noalign{\smallskip}
\multirow{2}*{Methods} & \multirow{2}*{Training Data} & \multicolumn{4}{c}{Regular} & \multicolumn{4}{c}{Irregular}  \\
\multirow{2}*{~} & \multirow{2}*{~} & IIIT5k & SVT & IC13 & Avg & SVTP & IC15 & CUTE & Avg \\
\noalign{\smallskip}
\hline
\noalign{\smallskip}
CRNN~\cite{shi2016end} & ST+MJ & 78.2 & 80.9 & 89.4 & 81.0 & - & - & - & - \\
ASTER~\cite{shi2018aster} & ST+MJ & 93.4 & 89.5 & 91.8 & 92.5 & 78.5 & 76.1 & 79.5 & 76.9 \\
TRBA~\cite{baek2019wrong} & ST+MJ & 87.9 & 87.5 & 92.3 & 88.8 & 79.2 & 77.6 & 74.0 & 77.6  \\
DAN~\cite{wang2020decoupled} & ST+MJ & 94.3 & 89.2 & 93.9 & 93.5 & 80.0 & 74.5 & 84.4 & 76.6 \\
NRTR~\cite{sheng2019nrtr} & ST+MJ & 90.1 & 91.5 & 95.8 & 91.5 & 86.6 & 79.4 & 80.9 & 81.1 \\
RobustScanner~\cite{yue2020robustscanner} & ST+MJ & 95.3 & 88.1 & 94.8 & 94.2 & 79.5 & 77.1 & 90.3 & 78.9  \\
SAR~\cite{li2019show} & ST+MJ & 91.5 & 84.5 & 91.0 & 90.4 & 76.4 & 69.2 & 83.3 & 72.1  \\
SEED~\cite{qiao2020seed} & ST+MJ & 93.8 & 89.6 & 92.8 & 93.0 & 81.4 & 80.0 & 83.6 & 80.6  \\
SCATTER~\cite{litman2020scatter} & ST+MJ & 93.2 & 90.9 & 94.1 & 93.1 & 86.2 & 82.0 & 84.8 & 83.2  \\
SATRN~\cite{lee2020recognizing} & ST+MJ & 92.8 & 91.3 & 94.1 & 92.9 & 86.5 & 79.0 & 87.8 & 81.5 \\
Text is Text~\cite{bhunia2021text} & ST+MJ & 92.3 & 89.9 & 93.3 & 92.2 & 84.4 & 76.9 & 86.3 & 79.4 \\
ABINet-LV~\cite{fang2021read} & ST+MJ+WiKi & {\bf 96.2} & 93.5 & {\bf 97.4} & {\bf 96.1} & 89.3 & \underline{86.0} & 89.2 & \underline{87.0} \\
S-GTR~\cite{he2021visual} & ST+MJ & 95.8 & \underline{94.1} & \underline{96.8} & \underline{95.8} & 87.9 & 84.6 & {\bf 92.3} & 86.0 \\
\noalign{\smallskip}
\hline
\noalign{\smallskip}
Baseline* & ST+MJ & 94.7 & 92.3 & 95.5 & 94.5 & 87.1 & 83.3 & 89.6 & 84.7  \\
Baseline + Corner & ST+MJ & 95.1 & \underline{94.1} & 95.7 & 95.1 & \underline{90.1} & 84.9 & 90.3 & 86.5  \\
Baseline + CC loss & ST+MJ & 95.4 & 92.0 & 96.1 & 95.1 & 88.2 & 83.9  & 89.8 & 85.4 \\
CornerTransformer & ST+MJ & \underline{95.9} & {\bf 94.6} & 96.4 & \underline{95.8} & {\bf 91.5} & {\bf 86.3} & \underline{92.0} & {\bf 88.0}  \\
\hline
\end{tabular}
}
\end{center}
\end{table}
\setlength{\tabcolsep}{1.4pt}

\subsection{Further Visualization and Analysis}

\begin{figure}
\centering
\includegraphics[width=0.85\textwidth]{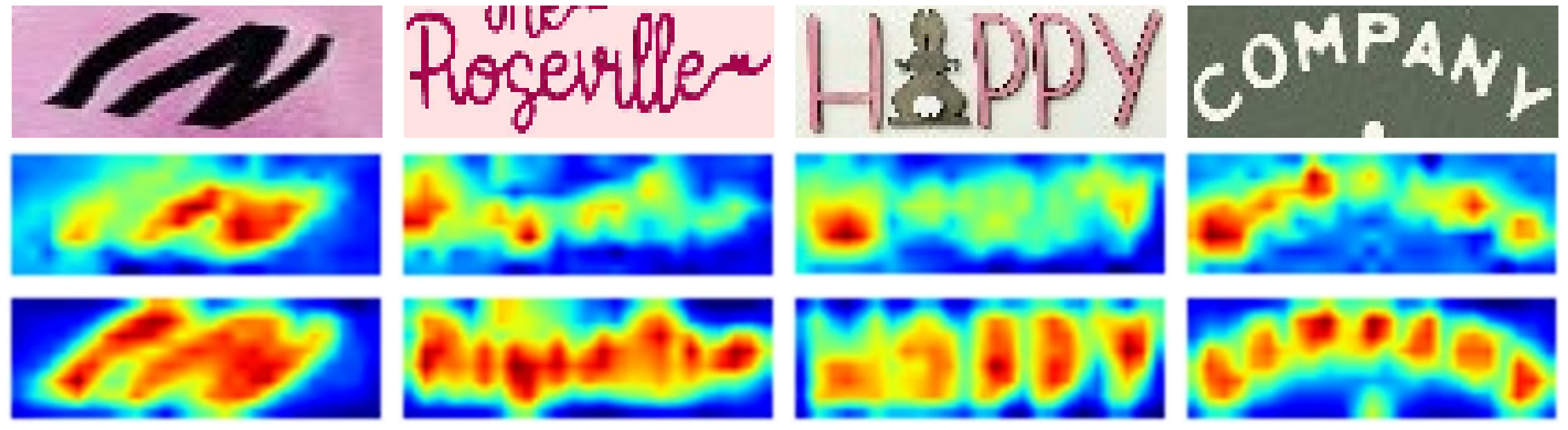}
\caption{Visualization for the feature map of the encoder output. First row: input images; Second row: feature maps of the baseline; Third row: feature maps of the baseline equipped with the corner-query cross-attention}
\label{fig:feature}
\end{figure}

\noindent\textbf{Corner directs more accurate attention.}
To intuitively verify the effectiveness of our corner-query cross-attention, exploring the essential mechanism why the corner map can improve the model performance, we visualize the feature map of the final output from our corner-guided encoder, as shown in Fig~\ref{fig:feature}. Evidently, for various text images with deformation, ligature, art design, and curve, our encoder can accurately focus on the position of each character, and there are apparent margins between characters. More importantly, our encoder can sometimes even focus on fine-grained features like character strokes, despite not providing any character-level or stroke-level annotations. All these good properties benefit from the corner-query cross-attention. The corner map contains the keypoints of the character strokes, and the corner-query attention enables the corner to seek the image features of interest (that is to seek other positions of the current character but not another character). Therefore, a corner point can gradually focus on the stroke feature up to the whole character feature. Besides, the corner map is very sparse and naturally separates each character. 

\begin{figure}
\centering
\subfigure[CE loss]{\includegraphics[width = 0.4\textwidth]{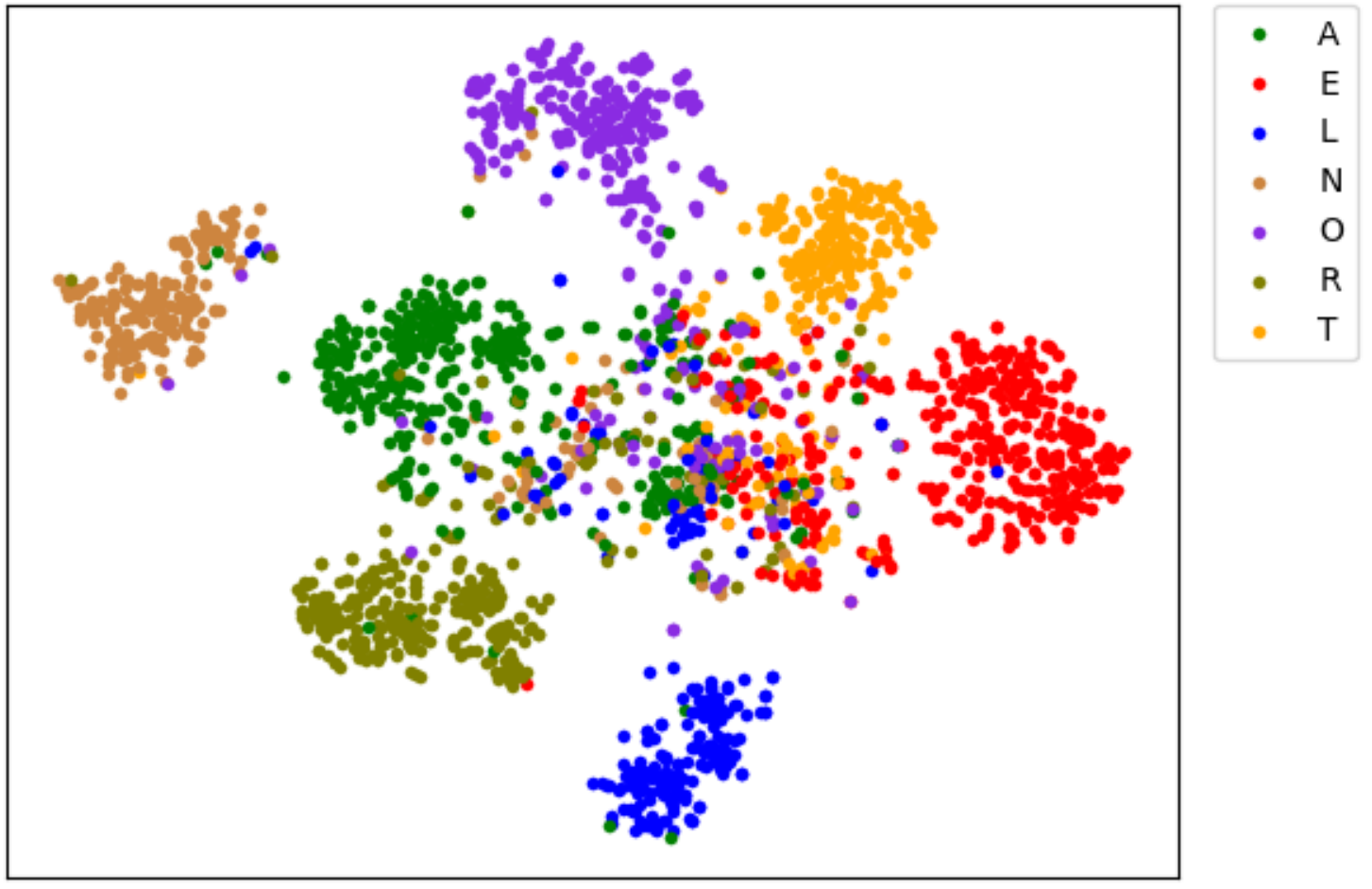}}
    \label{fig:short-a}
\subfigure[CE loss + CC loss ]{\includegraphics[width = 0.4\textwidth]{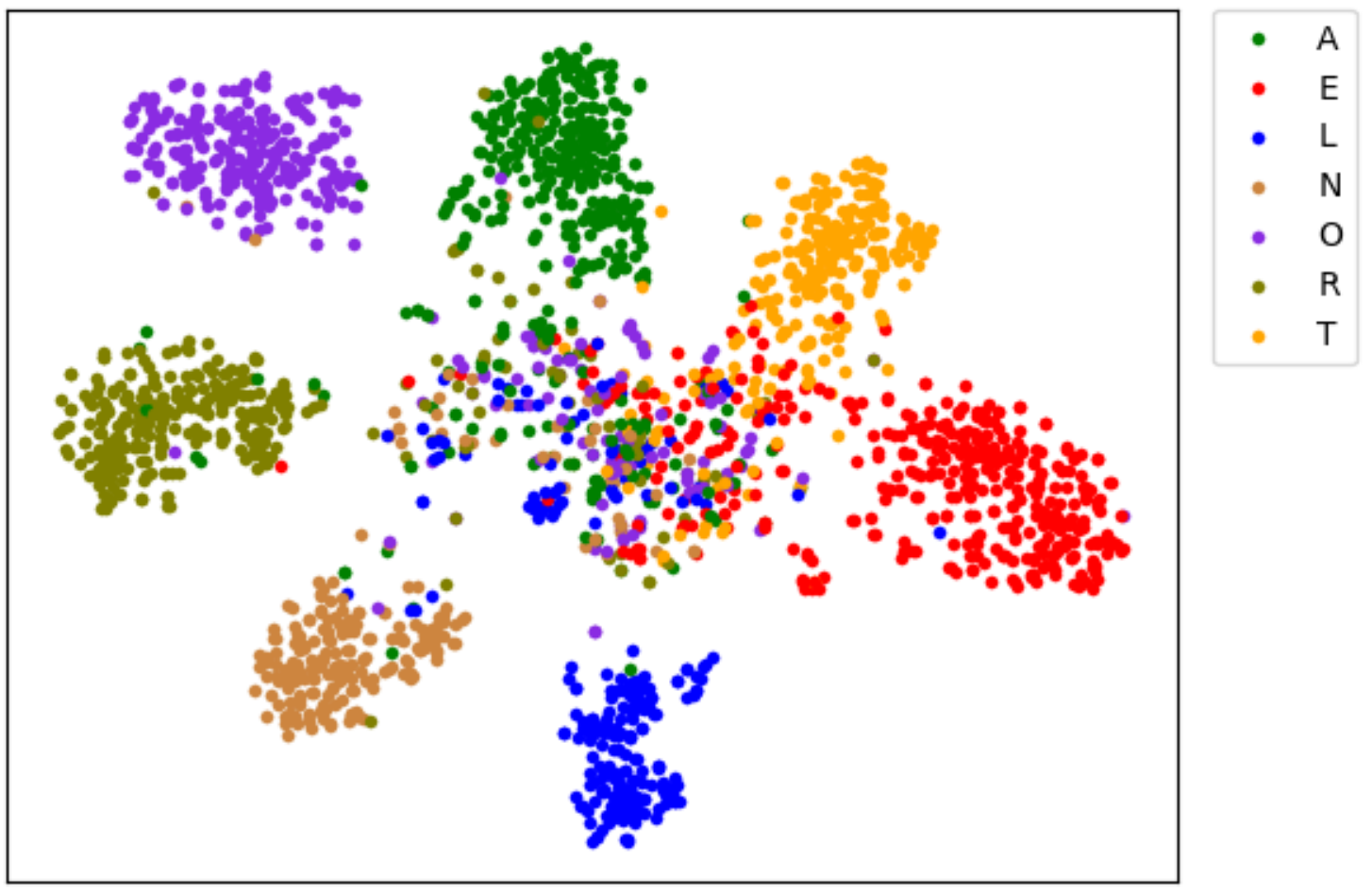}}
    \label{fig:short-b}
 
\caption{Visualization for the character feature distribution of the decoder output } 
\label{fig:loss}
\end{figure}
\noindent\textbf{Character contrastive loss improves class representation.}
In order to verify the effectiveness of our character contrastive loss and justify the motivation for designing this loss, we perform dimension reduction on the final output features of the CornerTransformer decoder and use t-SNE~\cite{van2008visualizing} to visualize the distribution of character features. Fig.~\ref{fig:loss} demonstrates the feature distributions of randomly selected 7 characters. Obviously, compared with the baseline using only the cross-entropy loss, when adding the character contrastive loss, the features of each character class are clustered together, and the features of different classes are far away from each other. This phenomenon is in line with our design that characters of the same category are positive samples and those of different categories are negative samples.

\subsection{Limitations}
\begin{figure}
\centering
\includegraphics[width=0.9\textwidth]{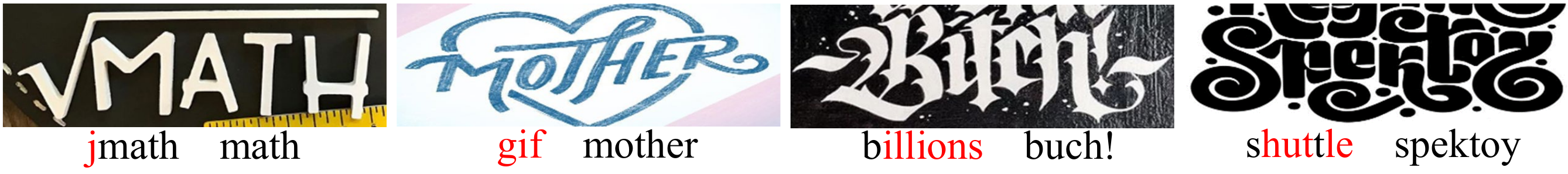}
\caption{Failure examples for artistic text recognition. Each image is along with our result and the ground truth}
\label{fig:failure}
\end{figure}

For some extremely difficult artistic texts, CornerTransformer may fail to achieve correct results. A few failure examples are shown in Fig.~\ref{fig:failure}. When decorative patterns from the background have exactly the same appearance and similar shape as the texts, it is difficult to distinguish whether these patterns belong to texts or not. These are indeed challenging examples for any text recognizer.

\vspace{-4ex}
\section{Conclusion}
In this paper, we focus on a new challenging task of artistic text recognition. To tackle the difficulties of this task, we introduce the corner point map as a robust representation for the artistic text image and present the corner-query cross-attention mechanism to make the model achieve more accurate attention. We also design a character contrastive loss to learn the invariant features of characters, leading to tight clustering of features. In order to benchmark the performance of different models, we provide the WordArt dataset. Experimental results demonstrate the remarkable superiority of our CornerTransformer on artistic text recognition. Interestingly, we achieve state-of-the-art results on several scene text datasets with small and blurred images. We hope the proposed WordArt dataset can encourage more advanced text recognition models, and the corner-based design can offer insights to other challenging recognition tasks.

\vspace{2ex}
\noindent\textbf{Acknowledgements}
This work was supported by the National Natural Science Foundation of China 61733007.

\clearpage
%
%
\bibliographystyle{splncs04}
\bibliography{egbib}
\end{document}